\documentclass[10pt,twocolumn,letterpaper]{article}

\usepackage{cvpr}              

\usepackage{graphicx}
\usepackage{amsmath}
\usepackage{amssymb}
\usepackage{booktabs}

\usepackage{microtype} 
\usepackage{multirow}
\usepackage{booktabs}
\usepackage[utf8]{inputenc} 
\usepackage{algorithm}
\usepackage{algpseudocode}
\usepackage{dblfloatfix}
\usepackage[accsupp]{axessibility}

\usepackage[pagebackref,breaklinks,colorlinks]{hyperref}

\usepackage[capitalize]{cleveref}
\crefname{section}{Sec.}{Secs.}
\Crefname{section}{Section}{Sections}
\Crefname{table}{Table}{Tables}
\crefname{table}{Tab.}{Tabs.}

\def\cvprPaperID{48} 
\def\confName{CVPR}
\def\confYear{2023}

\begin{document}

\title{LSFSL: Leveraging Shape Information in Few-shot Learning}

\author{Deepan Chakravarthi Padmanabhan\textsuperscript{\rm 1},
Shruthi Gowda\textsuperscript{\rm 1,2},
Elahe Arani\textsuperscript{\rm 1,2,}\thanks{Equal contribution.},
Bahram Zonooz\textsuperscript{\rm 1,2,*}\\
\textsuperscript{1}Advanced Research Lab, NavInfo Europe, Eindhoven, The Netherlands\\
\textsuperscript{2}Eindhoven University of Technology, Eindhoven, The Netherlands\\
{\tt\small deepangrad@gmail.com, shruthi.ngowda@outlook.com, \{e.arani, bahram.zonooz\}@gmai1.com}
}
\maketitle

\begin{abstract}

Few-shot learning (FSL) techniques seek to learn the underlying patterns in data using fewer samples, analogous to how humans learn from limited experience. In this limited-data scenario, the challenges associated with deep neural networks, such as shortcut learning and texture bias behaviors, are further exacerbated. Moreover, the significance of addressing shortcut learning is not yet fully explored in the few-shot setup. To address these issues, we propose LSFSL, which enforces the model to learn more generalizable features utilizing the implicit prior information present in the data. Through comprehensive analyses, we demonstrate that LSFSL-trained models are less vulnerable to alteration in color schemes, statistical correlations, and adversarial perturbations leveraging the global semantics in the data. Our findings highlight the potential of incorporating relevant priors in few-shot approaches to increase robustness and generalization.


\end{abstract}


\section{Introduction}
\label{section:intro}



Intelligent systems based on Deep Neural Networks (DNN) have achieved impressive performance in perception tasks \cite{Krizhevsky_ImageNet,Arani_od_survey,Taghanaki_sem_seg_survey}, yet DNNs remain limited in their ability to acquire new skills with little experience. Acquisition of novel skills with limited data is considered an important measure of intelligence \cite{chollet2019measure}, however, conventional supervised learning methods require significant amounts of labeled training data and computational resources. Therefore, in the quest for the building of intelligent agents, few-shot learning (FSL) \cite{Wang_FSLReview,Song_FSLReview} plays an imperative role. FSL is a learning paradigm that addresses these challenges by utilizing a small set of training samples to quickly learn the necessary skills and generalize to novel tasks with less direct supervision than conventional learning \cite{Juliani_consciousfn}.

Neural networks in the conventional training setup inadvertently learn spurious correlations in the data as shortcuts \cite{Geirhos_shortcut} and are biased to learn local texture information \cite{Geirhos_texturebias} instead of focusing on global features. These challenges are further accentuated in the FSL setup due to the limited training data \cite{Ringer_texturebiasfsl}, which greatly affects the generalization and robustness of the FSL models. Furthermore, the over-parameterized few-shot models lead to overfitting on the limited data. The significance of addressing the challenges of shortcut learning and texture bias in the FSL domain has not been thoroughly studied. Consequently, tackling these issues will aid in learning robust features that enhance the generalization of few-shot models in diverse real-world scenarios.


On the other hand, humans excel at learning with limited data to quickly adapt to new situations and remain both precise and robust in a constantly changing environment with minimal experience \cite{Lynn_dynamic_obj_recognition,Ayzenberg_skeletal_shape_descriptions}. This ability of humans to learn novel tasks with little to no prior experience or knowledge can be attributed to structural priors \cite{kuhl2000new}. For instance, the brain encompasses a structural prior to language acquisition that provides a meaningful disposition toward language learning in infants. 
Furthermore, studies on developmental cognitive neuroscience \cite{Elder_shape_importance,Marr_shape_importance} report the utility of semantic information by infants \cite{Landau_infantshapebias, Gil_infantshapebias} and adults \cite{Geirhos_texturebias} to recognize objects. Therefore, we leverage the prior knowledge already implicitly available in the data as a supervisory signal to help few-shot models learn the intended solution.

Inspired by cognitive biases, we propose an approach called \textit{Leveraging Shape Information in Few-shot Learning (LSFSL)}, which injects prior knowledge into the few-shot models in the form of shape awareness. This shape bias addresses the texture bias in the neural network and aids in learning robust global features. Consequently, the shape information might improve the generalization and offer a robust solution in FSL. The implicit shape information already presented in the visual data is used to train the few-shot models without the need for any additional dataset or generative techniques. During training, we learn a dual network architecture, where one model extracts and leverages shape information from standard RGB input. This shape information is distilled into the model trained with standard RGB image in synchrony through latent space and decision space alignment objectives. The alignment objectives regularize both networks and address overfitting to trivial solutions. Thus, LSFSL aims to address the vulnerability of few-shot models to shortcuts and texture bias by incorporating shape awareness. 

With extensive experiments on multiple datasets and settings, we show that our proposed approach boosts the generalization and robustness of FSL models to a greater extent. 
LSFSL-trained models are less susceptible to spurious correlations, texture bias, and adversarial perturbations.
The contributions of this work are as follows:
\begin{itemize}
\item We propose an approach to integrate FSL models with a shape-based inductive bias.
\item We evaluate the susceptibility of LSFSL-trained models and counterpart baseline models to shortcut learning and adversarial attacks.
\item Our approach offers flexibility to be plugged into meta \cite{Snell_PNFSL} and non-meta \cite{Rizve_IER_Distill} learning based few-shot approaches to improve their performance.
\end{itemize}


\section{Related Work}
\label{section:related_work}

\paragraph{Few-shot Learning:} 
FSL methods are broadly divided into three categories.
Optimization-based methods utilize meta-learning to enable the model to converge with a few training samples quickly. 
Methods identify a good model initialization \cite{Chelsea_MAML,Beaulieu_ANML}, optimizers for effective weight update \cite{Sachin_Opt} or both \cite{Sungyong_OptModel}.
Hallucination-based methods address data deficiency by extending the support set with similar samples using feature hallucination \cite{Hariharan_hallucinate} or data augmentation techniques. 
This converts the low-shot problem approximately to a standard classification problem. 
Metric-based methods learn an embedding model to map each image to a latent feature space \cite{Hou_prototype_classifiers}. 
The query samples are classified as the closest support category in the feature space.
Additionally, FSL methods incorporate meta-learning \cite{Chelsea_MAML, Snell_PNFSL} or non-meta standard supervised learning based \cite{Rizve_IER_Distill, Tian_RFS} training strategy to train a base learner.  
The resultant models from the later methods serve as an excellent embedding model and illustrate the ability to be used as a common architecture for different FSL settings.

\paragraph{Knowledge Distillation in FSL:}
With the aim of developing a good base embedding model to improve few-shot generalization, RFS-Distill \cite{Tian_RFS} incorporate sequential self-distillation after pretraining the model on base classes following Born-again Networks (BAN) \cite{Furlanello_BAN}.
This Born-again distillation with feature normalization enhances the representations from the backbones until a certain generation. 
Similarly, Invariant and Equivariant Representations (IER-Distill) \cite{Rizve_IER_Distill} utilize knowledge distillation to incorporate meaningful inductive biases into the base feature extractor.
IER incorporates both invariance and equivariance inductive biases of certain geometric transformations. 
Unlike RFS, IER employs a multi-head distillation over standard supervised distillation.
Liu \etal \cite{Liu_online_self_distillation} learn a base embedding model by unifying the two-stage training in RFS-Distill using online self-distillation.
We incorporate shape bias by distilling features in latent space and model decisions between two networks trained synchronously.

\paragraph{Overcoming Biases in FSL:}
The biases affecting the generalization of conventional DNNs create a larger impact on the FSL setup. 
Ringer \etal, study the effects of texture bias illustrated by Convolutional Neural Networks (CNN) in FSL \cite{Ringer_texturebiasfsl} and address the same by modifying the training data with a combination of non-texture and texture-based images. 
The image background also serves as a shortcut in the FSL scenario and impacts classification performance.
Luo \etal \cite{Luo_rectifybgshortcut} mitigates the problem by sampling the foreground and original images in the training and testing phases. 
Finally, even though the feature distribution of the base classes used in pretraining FSL models is well defined, the distribution is affected by the domain shift incurred by the novel classes. 
Tao \etal \cite{Tao_poweringfinetuning} propose Distribution Calibration Module (DCM) and Selected Sampling (SS) to mitigate class-agnostic and class-specific bias, respectively, in the meta-testing phase. 
However, LSFSL utilizes the implicit shape information in the training data with alignment objectives that aid in improving the generalization of few-shot models by learning global features.

\paragraph{Shape Bias in FSL:}
Stojanov \etal \cite{Stojanov_3dshapebias} illustrates the effectiveness of incorporating explicit shape bias for few-shot classification. 
The method learns a point cloud and image embedding on the base classes using 3D point cloud data and image data, respectively.
However, the evaluation includes the nearest centroid classifier on the average of both embeddings of the query images. 
Unlike Stojanov \etal \cite{Stojanov_3dshapebias}, LSFSL does not require any additional synthetic 3D object dataset for FSL and does not enforce any restrictions on the potential applications of incorporating shape bias.


\begin{figure*}[t]
\centering
\includegraphics[width=\linewidth]{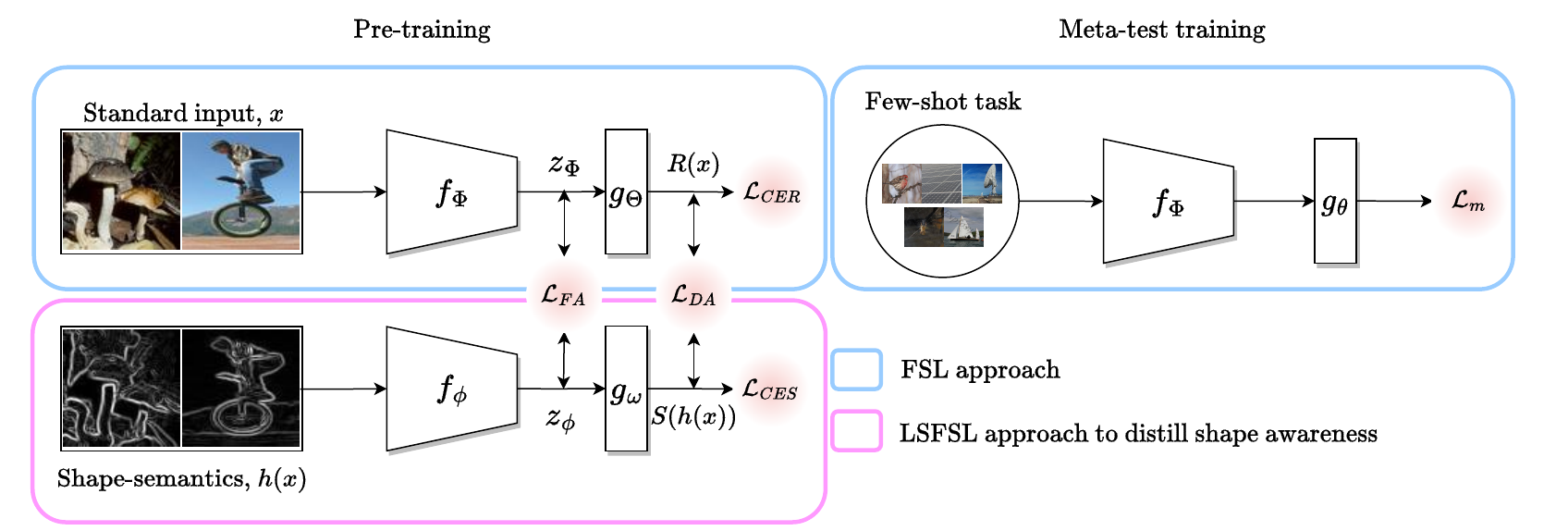}
\caption{LSFSL pipeline illustrating the pretraining and meta-test training pipeline. The pretraining stage incorporates the shape bias to address the few-shot model bias to texture. The shape semantics, $h(x)$, are obtained by applying a Sobel edge operator on the RGB input image, $x$. The meta-testing train stage fits a logistic regression on the RGB image features from the RIN model.}
\label{fig:v1}
\end{figure*}

\section{Methodology}
\label{section:methodology}

We formulate the FSL setup to train few-shot classifiers and then describe our proposed methodology to develop shape-aware few-shot classifiers.

\subsection{Preliminaries}

Generally, FSL for classification involves two steps: pretraining and meta-testing.
In the pretraining step, model $\mathcal{M}$ comprising a backbone feature extractor $f_\Phi$ and fully-connected layers $g_\Theta$ is trained on base classes $N_b$ from a base dataset $\mathcal{D}_b=\{(x_i, y_i)\}_{i=1}^B$. The classification loss function $L_{CER}$ and the parameter regularization term $\Re$ used to pretrain the model is given by,
\begin{equation}
\mathcal{M} =\underset{\Phi,\Theta}{\arg \min }\text{ } \mathbb{E}_{\mathcal{D}_b}\left[\mathcal{L}_{CER}\left(\mathcal{D}_{b} ; \Phi, \Theta\right)\right] + \Re(\Phi, \Theta)
\end{equation}

The pretraining strategy employs meta-learning \cite{Chelsea_MAML,Beaulieu_ANML, Snell_PNFSL} or standard supervised (non-meta) learning \cite{Tian_RFS,Rizve_IER_Distill} training setup. 
The meta-testing phase utilizes a dataset $\mathcal{D}_m$ with unseen novel classes to the classes in $\mathcal{D}_b$.
The meta-testing phase includes a series of few-shot tasks with data sampled from $\mathcal{D}_m$.
Each task contains a meta-test training and meta-test testing step using the support set $S$ = $(\mathcal{D}^{train}_i)_{i=1}^N$ and the query set $Q$ = ($\mathcal{D}^{test}_i)_{i=1}^N$, respectively. 
$N$ is the total number of tasks in the meta-testing phase.
Each support set contains $N_f$ unique novel classes and $k_f$ examples per class. 
Therefore, each few-shot task is represented as a $N_f$-way, $k_f$-shot task.

In the meta-test train phase, the pretrained model $\mathcal{M}$ acts as a fixed feature extractor utilizing $f_{\Phi}$ or is fine-tuned by updating specific layers or the entire model in meta-testing. 
The embeddings of $f_{\Phi}$ are used to train a simple classifier $g_{\theta}$ such as a logistic regression or a support vector machine \cite{Tian_RFS}.
Additionally, in certain cases, the embeddings are classified using non-parametric classifiers like nearest neighbor by estimating class prototypes \cite{Snell_PNFSL}. 
$\mathcal{L}_{m}$ is the loss function of the meta-test train phase with parameter regularization $\Re$. 
\begin{equation}
\Psi=\underset{\Psi}{\arg \min }\text{ } \mathbb{E}_{S}\left[\mathcal{L}_{m}\left(\mathcal{D}^{train} ; \Psi\right)\right] + \Re(\Psi)
\end{equation}
where $\Psi$ is given by,
\begin{equation}
\Psi= 
\begin{cases}
[\Phi^T, \theta^T]^T & \text { if both } \Phi \text{ and } \theta \text{ are trainable} \\
[\theta] & \text { if } \theta \text{ is trainable} \\
[\Phi] & \text { if } \Phi \text { is trainable}
\end{cases}
\end{equation}

The overall objective is to minimize the average test error over the distribution of the meta-test test set. 
The query set is sampled from the same distribution as the corresponding support set.
\begin{equation}
\mathbb{E}_{Q}\left[\mathcal{L}_{m}\left(\mathcal{D}^{test} ; \Psi \right)\right]
\end{equation}

\subsection{Incorporating Shape Awareness}

With the use of CNNs in majority of classification tasks \cite{Song_FSLReview}, the models are prone to learn local information \cite{Geirhos_shortcut}. This substantially affects generalization, even with the slightest perturbations, changes in background data, statistical irregularities, or color schemes.
This effect is further exacerbated in few-shot settings, as the distribution shift between the training and testing class is more prevalent \cite{Ringer_texturebiasfsl,Azad_texturebiasfss}.
This calls for a more robust model for few-shot learning.
Unlike CNNs, the human visual system is robust and recognizes objects under different conditions using fewer data. Studies have attributed this human behavior to cognitive biases of the brain or to gathered prior knowledge. The presence of cognitive biases helps to focus on the global discriminative shape features for recognition \cite{Landau_infantshapebias,Geirhos_texturebias}. Motivated by this, we propose to impart an additional bias in the form of a shape on top of the generic inductive biases of CNNs at the learning stage.

\begin{algorithm}[t]
\centering
\caption{LSFSL: Training Algorithm}
\label{algo:lsfsl_simple}
\begin{algorithmic}[1]
 \Statex \textbf{Input:} dataset $\mathcal{D}$, randomly initialized RIN model $R$ with feature extractor $f_{\Phi}$ and classifier $g_{\Theta}$, randomly initialized SIN model $S$ with feature extractor $f_{\phi}$ and classifier $g_{\omega}$, epochs $E$, softmax operator $\sigma$, stop-grad operator $SG$, cross-entropy loss $CE$, Kullback-Leibler divergence loss $KLD$, mean square error $MSE$, Sobel edge operator $h$, feature alignment loss factors ($\gamma_r$, $\gamma_r$) , decision alignment loss factors ($\lambda_r$, $\lambda_s$)
\For {epoch $e \in \{1, 2,..,E\}$}
  \State sample a mini-batch ${(x, y)} \sim \mathcal{D}$
  \State $x_{shape} = h(x)$
  \State $z_{\Phi}$ = $f_{\Phi}(x)$
  \State $z_{\phi}$ = $f_{\phi}(x_{shape})$
  \State $R(x)$ = $g_{\Theta} (f_{\Phi}(x))$
  \State $S(h(x))$ = $g_{\omega} (f_{\phi}(h(x_{shape})))$
  \State $\mathcal{L}_{CER}$ = $CE(\sigma(R(x)), y)$
  \State $\mathcal{L}_{CES}$ = $CE(\sigma(S(h(x))), y)$
  \State $\mathcal{L}_{FAR}$ = $MSE(z_{\Phi}, SG(z_{\phi}))$ \Comment{\small (Eq. \ref{equ:fa_rin})}
  \State $\mathcal{L}_{FAS}$ = $MSE(SG(z_{\Phi}), z_{\phi})$ \Comment{\small (Eq. \ref{equ:fa_sin})}
  \State $\mathcal{L_{FA}}$ = $\gamma_r \mathcal{L}_{FAR}$ + $\gamma_s \mathcal{L}_{FAS}$ \Comment{\small (Eq. \ref{equ:fa_mse})}
   \State $\mathcal{L}_{DAR}$ = $KLD(\sigma(R(x)))$, $SG(\sigma(S(h(x))))$ \Comment{\small (Eq. \ref{equ:da_rin})}
   \State $\mathcal{L}_{DAS}$ = $KLD(SG(\sigma(R(x)))$, $\sigma(S(h(x))))$ \Comment{\small (Eq. \ref{equ:da_sin})}
  \State $\mathcal{L_{DA}}$ = $\lambda_r \mathcal{L}_{DAR}$ + $\lambda_s \mathcal{L}_{DAS}$ \Comment{\small (Eq. \ref{equ:da_kl})}   
  \State $\mathcal{L}$ = $\mathcal{L}_{CER}$ + $\mathcal{L}_{CES}$ + $\mathcal{L}_{FA}$ + $\mathcal{L}_{DA}$ \Comment{\small (Eq. \ref{equ:loss_v1})}
  \State Update parameters of $R$ and $S$ based on $\mathcal{L}$ using Stochastic Gradient Descent (SGD)
\EndFor
\State \Return{RIN model $R$}
\end{algorithmic}
\end{algorithm}

We propose LSFSL, an approach to develop shape-aware FSL models leveraging the implicit shape information present in the RGB input image. 
Unlike \cite{Stojanov_3dshapebias}, we do not utilize additional datasets to develop shape-aware models. We incorporate the shape information into the model during the pretraining stage. Our LSFSL model synchronously trains two networks: RIN (standard RGB Input Network) and SIN (Shape Input Network), by distilling shape knowledge between the networks \cite{Shruthi_InBiaseD}. 
Each network comprises a backbone feature extractor followed by fully-connected layers for classification. The SIN network is fed with an image with enhanced shape semantics to extract the shape information pertaining to the data. 
To extract the shape information, we employ the Sobel edge operator \cite{Sobel_Sobel}, which identifies the shape information through discrete differentiation from the RGB input image in a computationally inexpensive way. 
As shown in Figure~\ref{fig:v1}, a standard input image $x$ and an image with shape semantics $h(x)$ are passed to the RIN and SIN networks, respectively. 
LSFSL employs two bias alignments to distill shape information from SIN to RIN.
The bias alignments: backbone feature alignment $\mathcal{L}_{FA}$ and decision alignment $\mathcal{L}_{DA}$ force RIN to focus on enhanced shape semantics.
This approach of learning the shape information along with the standard biases aids in achieving robust representation in the RIN network by aggregating information from RGB and the edge input image.


\begin{table*}[t]
\centering
\begin{tabular}{clcccccccccccc} 
\toprule
&\multicolumn{1}{c}{\multirow{2}{*}{Model}} & \multicolumn{2}{c}{miniImageNet} & \multicolumn{2}{c}{tieredImageNet} & \multicolumn{2}{c}{CIFAR-FS} & \multicolumn{2}{c}{FC100}\\ 
\cmidrule{3-4} \cmidrule{5-6} \cmidrule{7-8} \cmidrule{9-10}
&\multicolumn{1}{c}{} & 1-shot & 5-shot  & 1-shot & 5-shot & 1-shot & 5-shot  & 1-shot & 5-shot  \\ 
\midrule
\parbox[t]{2mm}{\multirow{4}{*}{\rotatebox[origin=c]{90}{Conv block}}}&MAML$^1$ \cite{Chelsea_MAML} & 48.70\tiny{$\pm$1.84}& 63.11\tiny{$\pm$0.92} & 51.67\tiny{$\pm$1.81} & 70.30\tiny{$\pm$1.75} & 58.90\tiny{$\pm$1.90}& 71.50\tiny{$\pm$1.00}& 38.10\tiny{$\pm$1.70} & 50.4\tiny{$\pm$1.00} \\
&PN$^2$ \cite{Snell_PNFSL} & 49.42\tiny{$\pm$0.78} & 68.20\tiny{$\pm$0.66} & 53.31\tiny{$\pm$0.89} & 72.69\tiny{$\pm$0.74} & 55.50\tiny{$\pm$0.0.70} & 72.00\tiny{$\pm$0.6} & 35.30\tiny{$\pm$0.60} & 48.60\tiny{$\pm$0.60} \\
&RN$^3$ \cite{Sung_RelationNet} & 50.44\tiny{$\pm$0.82} & 65.32\tiny{$\pm$0.70} & 54.48\tiny{$\pm$0.93} & 71.32\tiny{$\pm$0.78} & 55.00\tiny{$\pm$0.10} & 69.30\tiny{$\pm$0.80} & - & - \\
&R2D2$^4$ \cite{Bertinetto_R2D2} & 51.20\tiny{$\pm$0.60} & 68.80\tiny{$\pm$0.10} & - & - & 65.30\tiny{$\pm$0.20} & 79.40\tiny{$\pm$0.10} & - & - \\
\midrule
\parbox[t]{2mm}{\multirow{4}{*}{\rotatebox[origin=c]{90}{WRN-28-10}}}&Boosting \cite{Gidaris_Boosting} & 63.77\tiny{$\pm$0.45} & 80.70\tiny{$\pm$0.33} & 70.53\tiny{$\pm$0.51} & 84.98\tiny{$\pm$0.36} & 73.62\tiny{$\pm$0.31} & 86.05\tiny{$\pm$0.22} & - & -\\
&Fine-tuning \cite{Dhillon_finetuning} & 57.73\tiny{$\pm$0.62} & 78.17\tiny{$\pm$0.49} & 66.58\tiny{$\pm$0.70} & 85.55\tiny{$\pm$0.48} & 76.58\tiny{$\pm$0.68} & 85.79\tiny{$\pm$0.50} & 43.16\tiny{$\pm$0.59} & 57.57\tiny{$\pm$0.55} \\
&LEO${ }^{\dagger}$ \cite{Rusu_LEO}  & 61.76\tiny{$\pm$0.08} & 77.59\tiny{$\pm$0.12} & 66.33\tiny{$\pm$0.05} & 81.44\tiny{$\pm$0.09} & - & - & - & -\\
&AWGIM \cite{Guo_AWGIM} & 63.12\tiny{$\pm$0.08} & 78.40\tiny{$\pm$0.11} & 67.69\tiny{$\pm$0.11} & 82.82\tiny{$\pm$0.13} & - & - & - & -\\
\midrule
\parbox[t]{2mm}{\multirow{9}{*}{\rotatebox[origin=c]{90}{ResNet-12}}}&TEWAM \cite{Qiao_TEWAM} & 60.07\tiny{$\pm$na} & 75.90\tiny{$\pm$na} & - & - & 70.4\tiny{$\pm$na} & 81.30\tiny{$\pm$na} & - & - \\
&Shot-Free \cite{Ravichandran_Shotfree} & 59.04\tiny{$\pm$na} & 77.64\tiny{$\pm$na} & 63.52\tiny{$\pm$na} & 82.59\tiny{$\pm$na} & - & -& - & -\\
&MetaOptNet \cite{Lee_MetaOptNet} & 62.64\tiny{$\pm$0.61} & 78.63\tiny{$\pm$0.46} & 65.99\tiny{$\pm$0.72} & 81.56\tiny{$\pm$0.53} & 72.60\tiny{$\pm$0.70} & 84.30\tiny{$\pm$0.50}& 41.10\tiny{$\pm$0.60} & 55.50\tiny{$\pm$0.60} \\
&DSN-MR \cite{Simon_DSNMR} & 64.60\tiny{$\pm$0.72} & 79.51\tiny{$\pm$0.50} & 67.39\tiny{$\pm$0.82} & 82.85\tiny{$\pm$0.56} & - & - & - & -\\
&TADAM \cite{Oreshkin_TADAM} & 58.50\tiny{$\pm$0.3} & 76.70\tiny{$\pm$0.3} & - & - & - & - & 40.10\tiny{$\pm$0.40} & 56.10\tiny{$\pm$0.40} \\
&PN \cite{Luo_rectifybgshortcut} & 61.19\tiny{$\pm$0.40} & 76.50\tiny{$\pm$0.45} & - & - & - & - & - & - \\
&FSL \cite{Tian_RFS} & 62.02\tiny{$\pm$0.63} & 79.64\tiny{$\pm$0.44} & 69.74\tiny{$\pm$0.72} & 84.41\tiny{$\pm$0.55} & 71.50\tiny{$\pm$0.8} & 86.0\tiny{$\pm$0.5} & 42.60\tiny{$\pm$0.7} & 59.10\tiny{$\pm$0.6} \\ 
&LSFSL (Ours) & \textbf{64.67\tiny{$\pm$0.49}} & \textbf{81.79\tiny{$\pm$0.18}} & \textbf{71.17\tiny{$\pm$0.52}} & \textbf{86.23\tiny{$\pm$22}} & \textbf{73.45\tiny{$\pm$0.27}} & \textbf{87.07\tiny{$\pm$0.17}} & \textbf{43.60\tiny{$\pm$0.11}} & \textbf{60.12\tiny{$\pm$0.17}} \\
\bottomrule
\end{tabular}
\caption{5-way few-shot performance in FSL benchmark datasets. The backbone feature extractors are provided in the first column. The superscripts indicate models with convolutional blocks as feature extractors. $^1$four Conv blocks with 32 filters each, $^2$four Conv blocks with 64 filters each, $^3$four Conv blocks with 64-96-128-256 filters, and $^4$four Conv blocks with 96-192-384-512 filters. $^{\dagger}$is trained on train and validation splits. 
FSL is the baseline RFS model and LSFSL is the shape-distilled RFS model trained using our proposed approach.}
\label{tab:results_fsl}
\end{table*}

The feature embeddings from RIN backbone $f_{\Phi}$ and the SIN backbone $f_{\phi}$ for a batch of training images $x$ are represented as $z_{\Phi}$ and $z_{\phi}$, respectively. 
The Sobel-operated shape image is represented as $h(x)$. 
Bias alignments are bidirectional, as noticed, SIN requires certain texture information to improve generalization (Equation \ref{equ:fa_sin}) and vice versa distills shape knowledge to RIN (Equation \ref{equ:fa_rin}).
\begin{equation}
\mathcal{L}_{FAR}={\mathbb{E}}[\left\|z_{\Phi}-\text{stopgrad}(z_{\phi})\right\|_{2}^{2}]
\label{equ:fa_rin}
\end{equation}
\begin{equation}
\mathcal{L}_{FAS}={\mathbb{E}}[\left\|\text{stopgrad}(z_{\Phi})-z_{\phi}\right\|_{2}^{2}]
\label{equ:fa_sin}
\end{equation}

Therefore, putting Equation \ref{equ:fa_rin} and Equation \ref{equ:fa_sin} together, a strict alignment of the backbone features of RIN and SIN is accomplished using mean squared error (MSE) as follow,
\begin{equation}
\mathcal{L}_{FA}=\gamma_r \mathcal{L}_{FAR} + \gamma_s \mathcal{L}_{FAS}
\label{equ:fa_mse}
\end{equation}
where $\gamma_r$ and $\gamma_s$ control the influence of texture and shape on the final loss.
This alignment of backbone features ensures that the earlier stages of RIN are more shape-aware in the representation space.
Therefore, the feature alignment captures the generic feature representations unaffected by changes in the color schemes, perturbations, and backgrounds. 

The decision alignment, $\mathcal{L}_{DA}$, is used to align the decision boundary of RIN and SIN. 
Enhancing the decision boundary of RIN with SIN forces RIN to utilize shape information for classification. 
Hence, $\mathcal{L}_{DA}$ forces the RIN to be less susceptible to learning from the shortcut cues in the data, such as color schemes and background information.
The bi-directionality in decision alignment incorporates Kullback-Leibler divergence $\mathcal{D}_{KL}$ individually for each component, as given in Equation \ref{equ:da_rin} and Equation \ref{equ:da_sin}.
\begin{equation}
\mathcal{L}_{DAR}=\mathcal{D}_{KL}(\sigma(R(x)) \| \text{stopgrad}(\sigma(S(h(x))))
\label{equ:da_rin}
\end{equation}
\begin{equation}
\mathcal{L}_{DAS}=\mathcal{D}_{KL}(\text{stopgrad}(\sigma(R(x))) \| \sigma(S(h(x)))
\label{equ:da_sin}
\end{equation}
\begin{equation}
\mathcal{L}_{DA}=\lambda_r \mathcal{L}_{DAR} + \lambda_s \mathcal{L}_{DAS}
\label{equ:da_kl}
\end{equation}
where $R(x)$ and $S(h(x))$ are the output logits from RIN and SIN, respectively. 
$\sigma$ is $\operatorname{softmax}$ operator. 
$\lambda_r$ and $\lambda_s$ control the influence of distilling shape to RIN and texture to SIN, respectively.
Therefore, decision alignment reduces the vulnerability of RIN to learning from superficial cues.

$\mathcal{L}_{CER}$ and $\mathcal{L}_{CES}$ are the cross-entropy loss for classifying the input images by RIN and SIN, respectively. 
This standard supervision loss improves the generalization of both networks.
The overall loss for training shape-aware FSL model using LSFSL is given by,
\begin{equation}
\mathcal{L}=\mathcal{L}_{CER}+ \mathcal{L}_{CES}+ \mathcal{L}_{DA} + \mathcal{L}_{FA}
\label{equ:loss_v1}
\end{equation}

The training algorithm for the proposed LSFSL approach is provided in Algorithm~\ref{algo:lsfsl_simple}. 
The meta-testing phase utilizes only the shape-aware RIN model that aggregated shape information in addition to other inductive biases. 
The model trained by the aforementioned procedure can be enhanced using sequential knowledge distillation \cite{Tian_RFS} (Algorithm~\ref{algo:lsfsl_distill}) or online self-distillation \cite{Liu_online_self_distillation} (Algorithm~\ref{algo:lsfsl_online}). 
This type of training is a generic procedure and extendable to both non-meta (Section~\ref{section:results_rfs_lsfsl}) or meta-learning-based models (Section~\ref{section:results_pn_lsfsl}).


\section{Experimental Setup}
\label{section:experimental_setup}

We report results on the following FSL benchmark datasets: miniImageNet~\cite{Vinyals_MatchingNets}, tieredImageNet~\cite{Ren_MetaSSFSC}, CIFAR-FS~\cite{Bertinetto_CIFAR-FS}, and FC100~\cite{Oreshkin_TADAM}.

\textbf{miniImageNet} incorporates $100$ randomly sampled categories from ImageNet and $600$ images per category. 
The train, validation, and test splits of the categories are $64$, $16$, and $20$, respectively.

\textbf{tieredImageNet} encompasses $34$ super-classes spanning $608$ ImageNet classes, making it more challenging than miniImageNet. 
The train, validation, and test splits based on the super-class/class combinations are $20/351$, $6/97$, and $8/160$

\textbf{CIFAR-FS} is derived by randomly splitting CIFAR-100 into $64$, $16$, and $20$ classes for the train, validation, and test set, respectively.

\textbf{FC100} splits CIFAR-100 based on super-classes similar to tieredImageNet. The train, validation, and test splits are $60$, $20$, and $20$, respectively.

\begin{table*}[t]
\centering
\begin{tabular}{lccccccc}
\toprule
\multirow{2}{*}{Model} & \multirow{2}{*}{Shot} & \multirow{2}{*}{miniImageNet}  & \multicolumn{5}{c}{Tinted-miniImageNet} \\ 
\cmidrule(l){4-8}
& & & {Q} & {S} & {PT} & {PT+Q} & {PT+S} \\
\midrule
FSL & \multirow{2}{*}{1} & 62.02\tiny{$\pm$0.63} & 59.05\tiny{$\pm$0.74} & 55.50\tiny{$\pm$0.67} & 46.70\tiny{$\pm$0.27} & 40.18\tiny{$\pm$1.03} & 31.34\tiny{$\pm$0.06} \\ 
LSFSL&      &  \textbf{64.67}\tiny{$\pm$0.49} & \textbf{62.15}\tiny{$\pm$0.64} & \textbf{62.53}\tiny{$\pm$0.46} & \textbf{48.79}\tiny{$\pm$0.03} & \textbf{45.94}\tiny{$\pm$0.51} & \textbf{33.39}\tiny{$\pm$0.19} \\ 
\midrule
FSL & \multirow{2}{*}{5} & 79.64\tiny{$\pm$0.44} & 74.87\tiny{$\pm$0.09} & 68.78\tiny{$\pm$0.23} & 66.02\tiny{$\pm$0.25} & 50.47\tiny{$\pm$1.13} & 37.50\tiny{$\pm$0.44} \\ 
LSFSL  &      &  \textbf{81.79}\tiny{$\pm$0.18} & \textbf{77.91}\tiny{$\pm$0.26} & \textbf{79.24}\tiny{$\pm$0.16} & \textbf{66.34}\tiny{$\pm$0.46} & \textbf{58.51}\tiny{$\pm$0.46} & \textbf{44.99}\tiny{$\pm$0.50} \\ 
\bottomrule
\end{tabular}
\caption{Comparison of FSL baseline RFS model and LSFSL-trained counterparts on varying degrees of class-specific tinted-miniImageNet in the 5-way setting. 
The tints are applied on the query set (Q), support set (S), and only the base dataset in the pretraining stage (PT), pretraining and meta-test training stage (PT + S), pretraining and meta-test test set (PT+Q) in increasing order of presence of spurious correlations.}
\label{tab:results_tintedminiimagenet}
\end{table*}

We demonstrate the utility of shape awareness instilled by LSFSL by training a prominent FSL method, RFS \cite{Tian_RFS} in the LSFSL framework. 
In pretraining, the RGB images are passed to RIN after applying random crop, random horizontal flip with probability $0.5$, and color jitter with $0.4$ brightness factor, contrast factor, and saturation factor. 
The shape images are generated by applying the Sobel filter to the RGB image upsampled by $2$. 
This is followed by applying a random crop and a random horizontal flip corresponding to the RGB image to the shape input. 
The crop size for the CIFAR-100 derivatives is $32\times32$, and that for the ImageNet derivatives is $84\times84$. 
For experiments on miniImageNet, CIFAR-FS, and FC100, we train for $65$ epochs with $0.1$ learning rate decay factor at epoch $60$. 
For experiments on tieredImageNet, we train for $60$ epochs with $0.1$ learning rate decay after $30$, $40$, and $50$ epochs. 
All experiments use an SGD optimizer and a ResNet-12 backbone.
RFS baseline results are replicated using the implementation details from \cite{Tian_RFS}.

In the results and analyses sections followed, the FSL baseline model represents the RFS baseline results, and LSFSL represents LSFSL enhanced RFS results unless stated otherwise.
For LSFSL, the meta-testing train stage for each randomly sampled task involves training a logistic regression classifier using the support RGB image features from the pretrained and fixed backbone.
The meta-testing test reports the classification accuracy on $15$ query samples per class over $600$ randomly sampled tasks. 
The few-shot classification 5-way accuracies and standard deviations are reported across 3 runs for all experiments unless stated otherwise. 


\section{FSL Performance}
\label{section:results_rfs_lsfsl}
We evaluate the few-shot performance in $5$-way $1$-shot and $5$-way $5$-shot settings for all baseline methods and the LSFSL model in Table~\ref{tab:results_fsl}.
It can be seen that LSFSL, with shape semantics, demonstrates a performance gain of $1-2\%$ over the FSL baseline model as well as other state-of-the-art FSL methods across all datasets, including the more challenging FC100 and tieredImageNet datasets. 
FSL baseline model containing ResNet-12 backbone and trained by leveraging shape outperforms methods having WRN-28-10 backbone (3 times more parameters than ResNet-12). 
This indicates that distilling shape semantics to RIN by aligning with the SIN improves overall performance.
The shape semantics act as an additional supervision to RIN and improve generalization.

\begin{figure}[t]
\centering
\includegraphics[width=\linewidth]{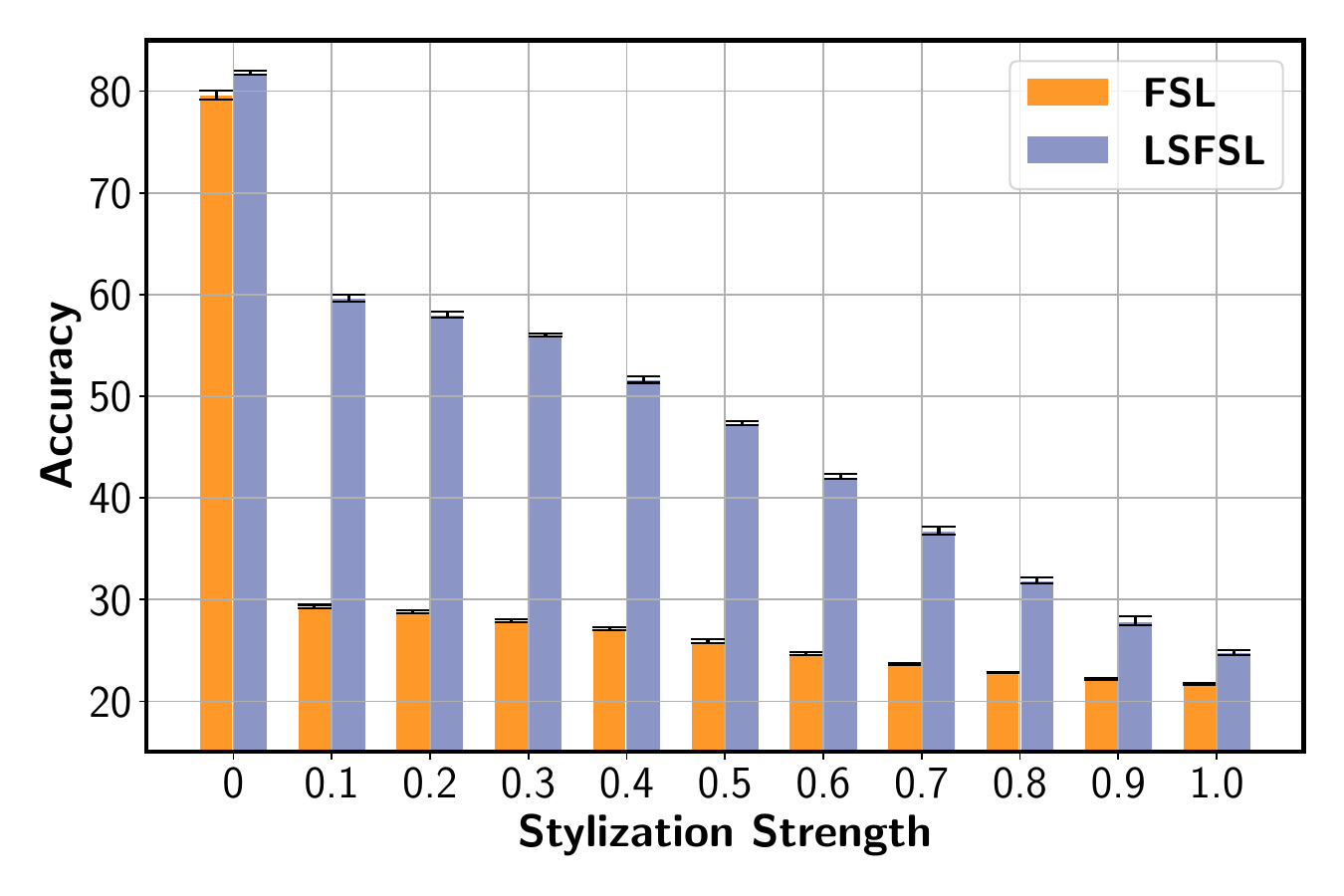}
\caption{Comparative evaluation of susceptibility of models pretrained and meta-test trained on miniImageNet images to local texture information in the 5-way setting. 
FSL is the baseline RFS model and LSFSL is the shape-distilled RFS model trained using our proposed approach.
}
\label{fig:results_stylizedminiimagenet}
\end{figure}

\section{Shortcut Learning}
\label{section:shortcut_learning}
Shortcut learning refers to the tendency of DNNs to learn trivial patterns in the data.
The decision rules learned in such a scenario substantially impede the generalization of DNNs to real-world scenarios~\cite{Geirhos_shortcut}.
These patterns can be of various forms: spurious correlations through tints~\cite{Jain_spuriouscorrelation}, statistical irregularities introduced through Fourier transforms~\cite{Jo_statisticalregularities}, and texture~\cite{Ringer_texturebiasfsl}. We look at the impact of learning shape semantics on reducing the susceptibility of DNNs to shortcut learning in FSL. All shortcut learning analyses are on miniImageNet unless mentioned otherwise.

\subsection{Texture Bias}
\label{section:texture_bias}
Unlike humans, DNNs strongly depend on local texture information to recognize objects~\cite{Geirhos_texturebias}. 
To test the impact of learning shape semantics on texture bias in FSL, we evaluate the models pretrained and meta-test trained with miniImageNet images on texture-biased query images (stylized-miniImageNet), generated following the protocol in~\cite{Huang_AdaIN}.
Specifically, we randomly select a texture from available texture pattern images and transfer the query image with texture transfer strength ranging between $0.1$ and $1.0$.
Figure~\ref{fig:results_stylizedminiimagenet} demonstrates that the shape semantics learned by the LSFSL framework help the network look beyond the local texture features and consistently outperform the baseline in all texture transfer strengths. 

\begin{figure}[t]
\centering
\includegraphics[width=\linewidth]{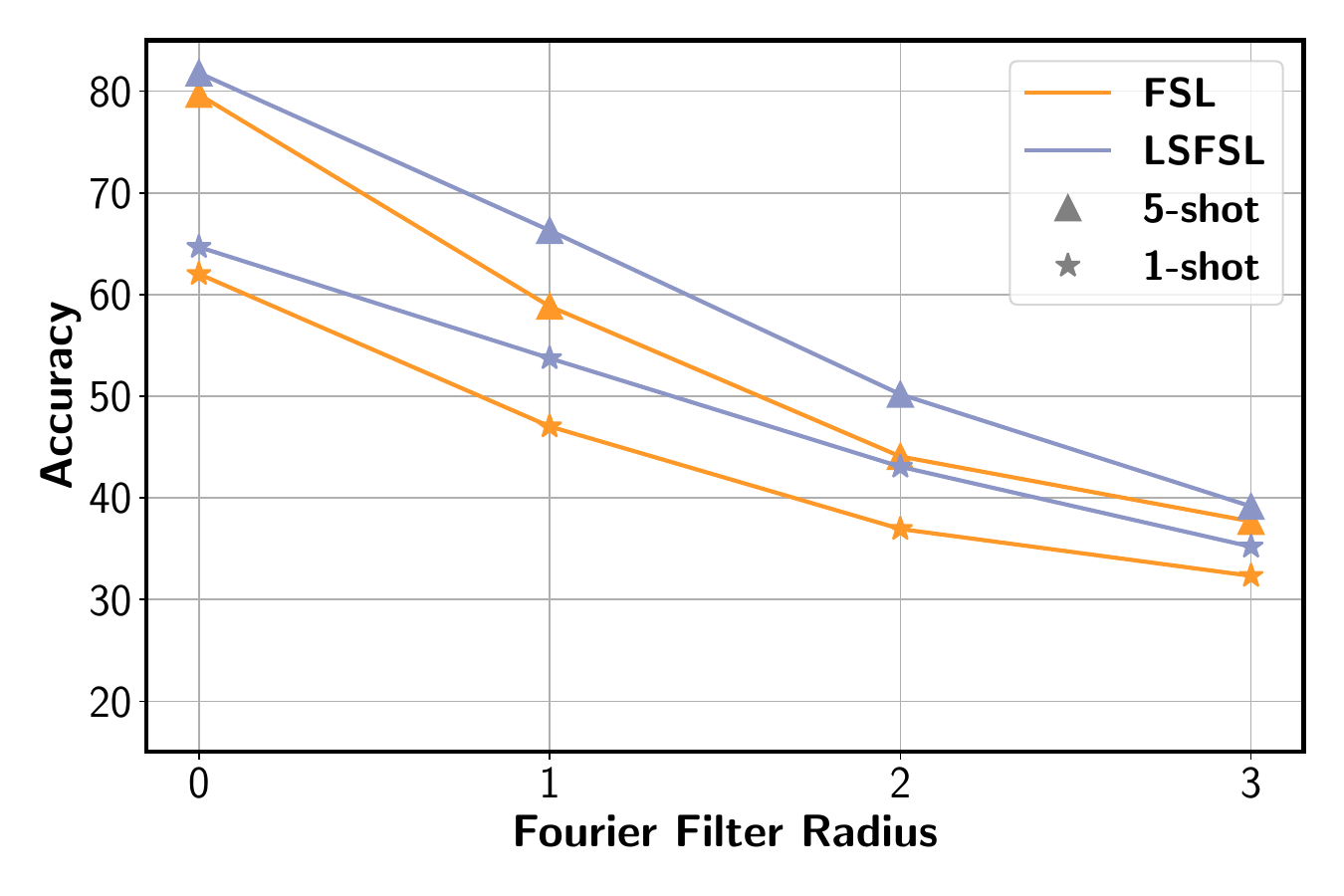}
\caption{Comparative evaluation of susceptibility to statistical regularities of models pretrained and meta-test trained on miniImageNet images in the 5-way setting.
The meta-test testing is performed on the Fourier low-pass filtered query images of varying filter radius in the 5-way setting.
FSL is the RFS model and LSFSL is the shape-distilled RFS model trained using our proposed approach.}
\label{fig:results_ftminiimagenet}
\end{figure}

\subsection{Spurious Correlation Analysis}
\label{section:spurious_correlation_analysis}
Training data collected in real-world scenarios include various inadvertent superficial cues.
DNNs tend to learn the class correlation with these cues and therefore illustrate a limited generalization~\cite{Geirhos_shortcut}.
We test the ability of LSFSL models trained with shape awareness to resist learning spurious correlations by analyzing the generalization across 5 different scenarios of tints.
Table~\ref{tab:results_tintedminiimagenet} illustrates performance under increasing levels of spurious correlations introduced through class-specific tints, arranged from left to right in ascending order.
We observe that the shape-aware LSFSL model significantly outperforms the corresponding baseline without shape awareness in all scenarios, even in the extreme scenarios of spurious correlations, i.e., PT+Q and PT+S. 
Thus, the shape-aware LSFSL-trained models are less susceptible to spurious correlations in the data. 


\subsection{Statistical Regularity Analysis}
\label{section:statistical_regularity_analysis}
Humans rely on high-level abstractions and structure in the data to recognize objects, whereas DNNs tend to learn statistical regularities that affect generalization in challenging testing scenarios~\cite{Jo_statisticalregularities}. 
We test the impact of distilling shape as an inductive bias on susceptibility to statistical regularities by evaluating query images after applying a radial low-pass Fourier filter at increasing severities. 
This filter preserves the visible high-level abstractions for human recognizability, but introduces a superficial statistical regularity by removing high-frequency components. 
Visualizations of the Fourier transformed images can be found in Appendix~\ref{section:app_visualizations}.
Figure~\ref{fig:results_ftminiimagenet} shows that shape-aware LSFSL models are less susceptible to statistical regularities than baseline models that lack shape awareness.
We contend that the shape awareness introduced by LSFSL models confronts the vulnerability of DNNs to learn shortcuts and takes a positive step towards the human perception of processing high-level and structural information to recognize objects.
Therefore, LSFSL leverages shape semantics to reduce the vulnerability of DNNs to shortcut cues such as texture, spurious correlations, and statistical regularities in FSL.

\begin{figure}[t]
\centering
\includegraphics[width=\linewidth]{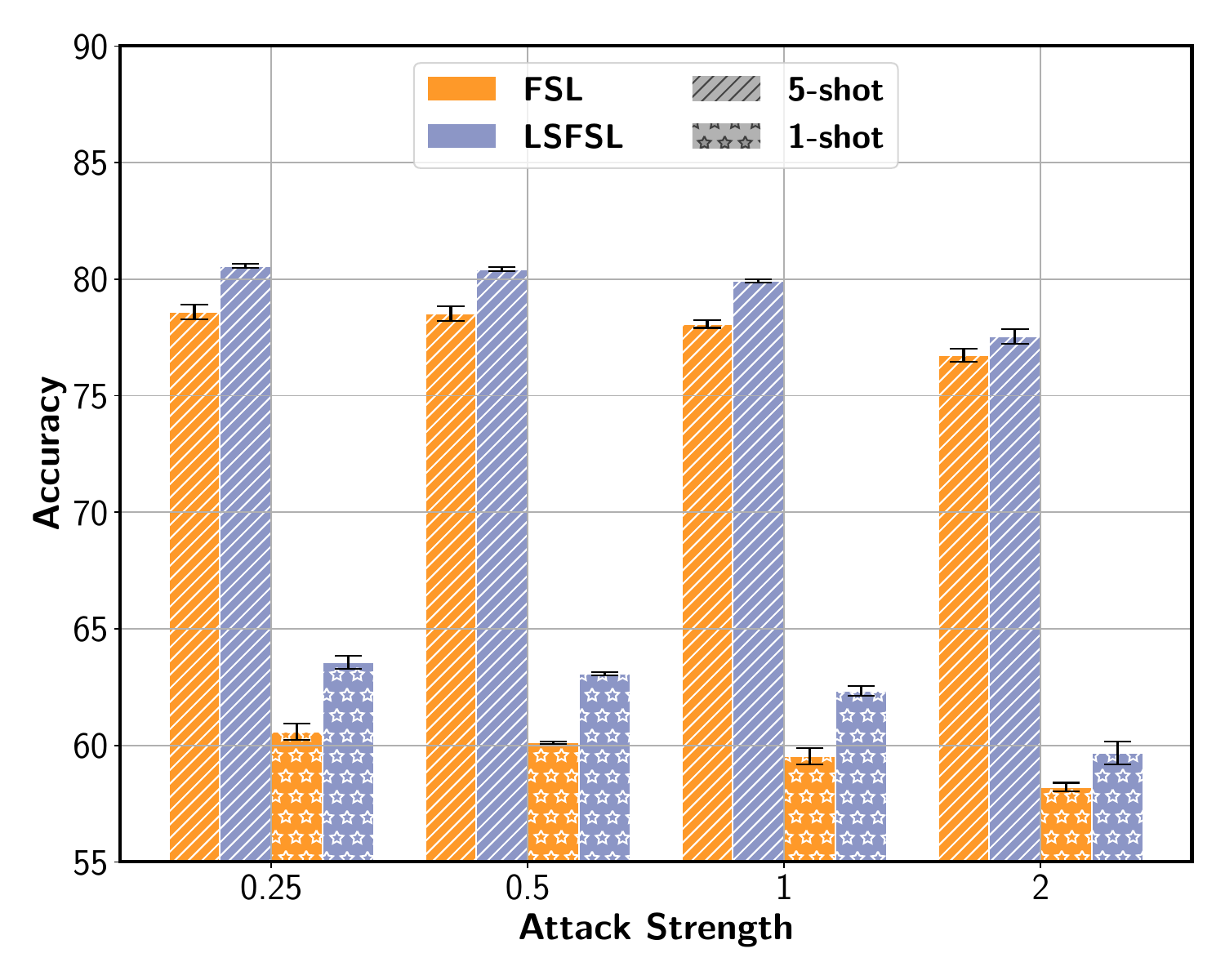}
\caption{Comparative evaluation of few-shot performance in the 5-way setting on adversarial query images of miniImageNet generated using Square attack at multiple strengths. FSL and LSFSL are the baseline and shape-aware RFS models trained using our proposed approach, respectively.}
\label{fig:results_square_attack}
\end{figure}


\section{Robustness}

Perturbations imperceptible to humans to input images result in adversarial examples that affect the generalization of DNNs and compromise the integrity of intelligent systems in safety-critical real-time applications. This complicates developing robust classifiers using only a few training samples, a scenario desired by many users~\cite{Subramanya_FSLAdversarialRobustness}. 
In contrast to DNNs, humans are robust to adversarial examples~\cite{Reddy_humanrobustness} and also learn to adapt quickly to new tasks with limited samples.
This is attributed to cognitive biases in the brain~\cite{Orhan_humanrobustness}. 
We study the utility of distilling shape semantics as an inductive bias in generalizing to adversarial examples. The query images are perturbed using a complex black box, Square~\cite{Andriushchenko_square} attack at multiple perturbation strengths. 
Figure~\ref{fig:results_square_attack} illustrates that LSFSL-trained models are more robust to adversarial query images. 
Therefore, the shape-aware counterparts of the baseline FSL methods are less prone to adversarial attacks due to learning high-level shape semantics information.

\begin{table*}[t]
\centering
\begin{tabular}{lcccccccccc} 
\toprule
$\mathcal{L_{CER}}$ & $\mathcal{L_{CES}}$ & $\mathcal{L_{DAR}}$ & $\mathcal{L_{DAS}}$ & $\mathcal{L_{FAR}}$ & $\mathcal{L_{FAS}}$ & 1-shot & 5-shot \\ 
\midrule
\checkmark & & & & & & 62.02\tiny{$\pm$0.63} & 79.64\tiny{$\pm$0.44} \\
\checkmark & \checkmark & \checkmark & & & & 63.86\tiny{$\pm$0.54} & 81.07\tiny{$\pm$0.29}\\
\checkmark & \checkmark & & \checkmark & & &  62.56\tiny{$\pm$0.51} & 80.72\tiny{$\pm$0.26} \\
\checkmark & \checkmark & \checkmark & \checkmark & & & 63.90\tiny{$\pm$0.55} & 81.25\tiny{$\pm$0.19} \\
\checkmark & \checkmark & & & \checkmark & & 63.03\tiny{$\pm$0.48} & 80.86\tiny{$\pm$0.29} \\
\checkmark & \checkmark & & & & \checkmark & 62.89\tiny{$\pm$0.72} & 81.00\tiny{$\pm$0.25} \\
\checkmark & \checkmark & & & \checkmark & \checkmark & 63.44\tiny{$\pm$0.59} & 81.24\tiny{$\pm$0.15} \\
\checkmark & \checkmark & \checkmark & \checkmark & \checkmark & \checkmark & \textbf{64.67}\tiny{$\pm$0.49} & \textbf{81.79}\tiny{$\pm$0.18} \\
\bottomrule
\end{tabular}
\caption{Ablation study to illustrate the effect of each component of the decision and backbone feature alignments. The study is performed using the miniImageNet dataset in a 5-way setting. RFS baseline FSL model is trained using the proposed LSFSL approach.}
\label{tab:ablation_eachalignment}
\end{table*}

\begin{table}[t]
\centering
\begin{tabular}{lcccccccc} 
\toprule
\multirow{2}{*}{Model} & \multicolumn{2}{c}{miniImageNet}  & \multicolumn{2}{c}{Tinted-miniImageNet} \\ 
\cmidrule{2-3} \cmidrule{4-5}
 &1-shot & 5-shot &1-shot & 5-shot \\ 
\midrule
FSL & 61.19\tiny{$\pm$0.40} & 76.50\tiny{$\pm$0.45} & 25.92\tiny{$\pm$0.24} & 28.67\tiny{$\pm$0.14} \\
LSFSL & \textbf{62.44}\tiny{$\pm$0.77} & \textbf{77.62}\tiny{$\pm$0.48} & \textbf{29.49}\tiny{$\pm$0.18} & \textbf{32.35}\tiny{$\pm$0.25} \\
\bottomrule
\end{tabular}
\caption{
Comparative evaluation of few-shot performance between conventional PN (indicated as FSL) and PN with shape bias distilled using LSFSL. 
The results are reported on 5-way setting. 
tinted-miniImageNet provided below incorporates tints in the query samples whereas the support images are without tint. }
\label{tab:results_pn_lsfsl}
\end{table}

\section{Ablation Study}

Earlier, we saw that instilling shape bias into the FSL models brings discernible benefits. 
We investigate the impact of each component of the LSFSL framework individually.
Table~\ref{tab:ablation_eachalignment} reports the contribution of each alignment term in Equations~\labelcref{equ:fa_rin,equ:fa_sin,equ:da_rin,equ:da_sin} to FSL performance. 
Each alignment improves the accuracy by a considerable amount. 
However, the decision and backbone alignment provides a significant few-shot performance boost. 
Therefore, in addition to improving FSL performance, the alignment objectives help reduce susceptibility to shortcut cues such as texture bias, spurious correlations, and statistical regularities.


\section{Generalization to Meta-learning}
\label{section:results_pn_lsfsl}

We illustrated earlier that incorporating shape information using LSFSL increases the generalization and robustness in various few-shot settings over RFS, a non-meta learning-based training setup. 
To evaluate LSFSL utility with meta-learning-based approaches, we considered adding LSFSL to the frequently reported Prototypical Networks (PN)~\cite{Snell_PNFSL} FSL approach.
The model is trained for 120 epochs with a learning rate of 0.05 and 1000 tasks per training epoch. 
The meta-test train estimates the prototypes from the support set image features and classifies using the nearest class prototype in the meta-test testing step. 
The meta-test test accuracy is reported as an average across 5 runs. 
Each run incorporates 2000 tasks with 15 query images per class in a task.
The FSL performance is reported in Table~\ref{tab:results_pn_lsfsl}. 
It is clear that we improve performance and reduce the susceptibility to learning shortcuts in the form of color schemes and texture. 
Thus, it is evident that our approach can be extended to both meta-learning and non-meta-learning-based FSL settings.


\section{Conclusion}
We propose an approach, LSFSL, that incorporates one of the cognitive biases to address the susceptibility of few-shot models to spurious cues and texture information. 
This vulnerability is particularly detrimental in the FSL paradigm compared to conventional learning scenarios.
Our shape-aware few-shot classifiers outperform the state-of-the-art methods across multiple FSL benchmarks in both $5$-way $5$-shot as well as $5$-way $1$-shot classification tasks. 
Additionally, the LSFSL approach demonstrates improved robustness against adversarial attacks and reduced sensitivity to shortcut cues, such as local texture information, spurious correlations, and statistical irregularities. 
Integrating shape semantics as an inductive bias leads to learning higher-level abstractions, thereby promoting both generalization and robustness. 
Lastly, we demonstrate the compatibility of our pipeline with both meta and non-meta-based FSL approaches. 
Our work aims to shed light on the untapped capabilities of DNN-based classifiers in low-data settings that are practical for real-world use, and we hope it opens up avenues for further research in this direction.




{\small
\bibliographystyle{ieee_fullname}
\bibliography{cvpr2023/lsfsl_cvpr2023}

\begin{thebibliography}{10}\itemsep=-1pt

\bibitem{Andriushchenko_square}
Maksym Andriushchenko, Francesco Croce, Nicolas Flammarion, and Matthias Hein.
\newblock Square attack: {A} query-efficient black-box adversarial attack via
  random search.
\newblock In Andrea Vedaldi, Horst Bischof, Thomas Brox, and Jan{-}Michael
  Frahm, editors, {\em Computer Vision - {ECCV} 2020 - 16th European
  Conference, Glasgow, UK, August 23-28, 2020, Proceedings, Part {XXIII}},
  volume 12368 of {\em Lecture Notes in Computer Science}, pages 484--501.
  Springer, 2020.

\bibitem{Arani_od_survey}
Elahe Arani, Shruthi Gowda, Ratnajit Mukherjee, Omar Magdy,
  Senthilkumar~Sockalingam Kathiresan, and Bahram Zonooz.
\newblock {A Comprehensive Study of Real-Time Object Detection Networks Across
  Multiple Domains: A Survey}.
\newblock {\em {Transactions on Machine Learning Research}}, 2022.
\newblock Survey Certification.

\bibitem{Ayzenberg_skeletal_shape_descriptions}
Vladislav Ayzenberg.
\newblock Skeletal descriptions of shape provide unique perceptual information
  for object recognition.
\newblock {\em Scientific Reports}, 9:9359, 06 2019.

\bibitem{Azad_texturebiasfss}
Reza Azad, Abdur~R. Fayjie, Claude Kauffmann, Ismail~Ben Ayed, Marco Pedersoli,
  and Jose Dolz.
\newblock {On the Texture Bias for Few-Shot {CNN} Segmentation}.
\newblock In {\em {{IEEE} Winter Conference on Applications of Computer Vision
  ({WACV})}}, pages 2673--2682. {IEEE}, 2021.

\bibitem{Sungyong_OptModel}
Sungyong Baik, Myungsub Choi, Janghoon Choi, Heewon Kim, and Kyoung~Mu Lee.
\newblock {Meta-Learning with Adaptive Hyperparameters}.
\newblock In Hugo Larochelle, Marc'Aurelio Ranzato, Raia Hadsell,
  Maria{-}Florina Balcan, and Hsuan{-}Tien Lin, editors, {\em Neural
  Information Processing Systems ({NeurIPS})}, 2020.

\bibitem{Beaulieu_ANML}
Shawn Beaulieu, Lapo Frati, Thomas Miconi, Joel Lehman, Kenneth~O. Stanley,
  Jeff Clune, and Nick Cheney.
\newblock {Learning to Continually Learn}.
\newblock In Giuseppe~De Giacomo, Alejandro Catal{\'{a}}, Bistra Dilkina,
  Michela Milano, Sen{\'{e}}n Barro, Alberto Bugar{\'{\i}}n, and
  J{\'{e}}r{\^{o}}me Lang, editors, {\em European Conference on Artificial
  Intelligence {(ECAI 2020)} - Conference on Prestigious Applications of
  Artificial Intelligence {(PAIS 2020)}}, volume 325 of {\em Frontiers in
  Artificial Intelligence and Applications}, pages 992--1001. {IOS} Press,
  2020.

\bibitem{Bertinetto_R2D2}
Luca Bertinetto, Jo{\~{a}}o~F. Henriques, Philip H.~S. Torr, and Andrea
  Vedaldi.
\newblock {Meta-learning with differentiable closed-form solvers}.
\newblock In {\em {International Conference on Learning Representations
  ({ICLR})}}. OpenReview.net, 2019.

\bibitem{Bertinetto_CIFAR-FS}
Luca Bertinetto, Jo{\~{a}}o~F. Henriques, Philip H.~S. Torr, and Andrea
  Vedaldi.
\newblock {Meta-learning with differentiable closed-form solvers}.
\newblock In {\em 7th International Conference on Learning Representations,
  {ICLR} 2019, New Orleans, LA, USA, May 6-9, 2019}. OpenReview.net, 2019.

\bibitem{chollet2019measure}
Fran{\c{c}}ois Chollet.
\newblock On the measure of intelligence.
\newblock {\em CoRR}, abs/1911.01547, 2019.

\bibitem{Dhillon_finetuning}
Guneet~Singh Dhillon, Pratik Chaudhari, Avinash Ravichandran, and Stefano
  Soatto.
\newblock {A Baseline for Few-Shot Image Classification}.
\newblock In {\em International Conference on Learning Representations
  ({ICLR})}. OpenReview.net, 2020.

\bibitem{Gil_infantshapebias}
Gil Diesendruck and Paul Bloom.
\newblock How specific is the shape bias?
\newblock {\em Child development}, 74:168--78, 2003.

\bibitem{Elder_shape_importance}
James~H. Elder.
\newblock Shape from contour: Computation and representation.
\newblock {\em Annual Review of Vision Science}, 4(1):423--450, 2018.
\newblock PMID: 30222530.

\bibitem{Chelsea_MAML}
Chelsea Finn, Pieter Abbeel, and Sergey Levine.
\newblock {Model-Agnostic Meta-Learning for Fast Adaptation of Deep Networks}.
\newblock In Doina Precup and Yee~Whye Teh, editors, {\em International
  Conference on Machine Learning ({ICML})}, volume~70 of {\em Proceedings of
  Machine Learning Research}, pages 1126--1135. {PMLR}, 2017.

\bibitem{Furlanello_BAN}
Tommaso Furlanello, Zachary~Chase Lipton, Michael Tschannen, Laurent Itti, and
  Anima Anandkumar.
\newblock {Born-Again Neural Networks}.
\newblock In Jennifer~G. Dy and Andreas Krause, editors, {\em {International
  Conference on Machine Learning {ICML}}}, volume~80 of {\em {Proceedings of
  Machine Learning Research}}, pages 1602--1611. {PMLR}, 2018.

\bibitem{Geirhos_shortcut}
Robert Geirhos, J{\"{o}}rn{-}Henrik Jacobsen, Claudio Michaelis, Richard~S.
  Zemel, Wieland Brendel, Matthias Bethge, and Felix~A. Wichmann.
\newblock {Shortcut learning in deep neural networks}.
\newblock {\em {Nature Machine Intelligence}}, 2(11):665--673, 2020.

\bibitem{Geirhos_texturebias}
Robert Geirhos, Patricia Rubisch, Claudio Michaelis, Matthias Bethge, Felix~A.
  Wichmann, and Wieland Brendel.
\newblock {ImageNet-trained CNNs are biased towards texture; increasing shape
  bias improves accuracy and robustness}.
\newblock In {\em {International Conference on Learning Representations
  ({ICLR})}}. OpenReview.net, 2019.

\bibitem{Gidaris_Boosting}
Spyros Gidaris, Andrei Bursuc, Nikos Komodakis, Patrick P{\'{e}}rez, and
  Matthieu Cord.
\newblock {Boosting Few-Shot Visual Learning With Self-Supervision}.
\newblock In {\em {{IEEE/CVF} International Conference on Computer Vision
  (ICCV)}}, pages 8058--8067. {IEEE}, 2019.

\bibitem{Shruthi_InBiaseD}
Shruthi Gowda, Bahram Zonooz, and Elahe Arani.
\newblock Inbiased: Inductive bias distillation to improve generalization and
  robustness through shape-awareness.
\newblock {\em CoRR}, abs/2206.05846, 2022.

\bibitem{Guo_AWGIM}
Yiluan Guo and Ngai{-}Man Cheung.
\newblock {Attentive Weights Generation for Few Shot Learning via Information
  Maximization}.
\newblock In {\em {{IEEE/CVF} Conference on Computer Vision and Pattern
  Recognition, ({CVPR})}}, pages 13496--13505. Computer Vision Foundation /
  {IEEE}, 2020.

\bibitem{Hariharan_hallucinate}
Bharath Hariharan and Ross~B. Girshick.
\newblock {Low-Shot Visual Recognition by Shrinking and Hallucinating
  Features}.
\newblock In {\em {IEEE} International Conference on Computer Vision ({ICCV})},
  pages 3037--3046. {IEEE} Computer Society, 2017.

\bibitem{Hou_prototype_classifiers}
Mingcheng Hou and Issei Sato.
\newblock A closer look at prototype classifier for few-shot image
  classification.
\newblock {\em CoRR}, abs/2110.05076, 2021.

\bibitem{Huang_AdaIN}
Xun Huang and Serge~J. Belongie.
\newblock Arbitrary style transfer in real-time with adaptive instance
  normalization.
\newblock {\em CoRR}, abs/1703.06868, 2017.

\bibitem{Jain_spuriouscorrelation}
Saachi Jain, Dimitris Tsipras, and Aleksander Madry.
\newblock Combining diverse feature priors.
\newblock In Kamalika Chaudhuri, Stefanie Jegelka, Le Song, Csaba
  Szepesv{\'{a}}ri, Gang Niu, and Sivan Sabato, editors, {\em International
  Conference on Machine Learning, {ICML} 2022, 17-23 July 2022, Baltimore,
  Maryland, {USA}}, volume 162 of {\em Proceedings of Machine Learning
  Research}, pages 9802--9832. {PMLR}, 2022.

\bibitem{Jo_statisticalregularities}
Jason Jo and Yoshua Bengio.
\newblock Measuring the tendency of cnns to learn surface statistical
  regularities.
\newblock {\em CoRR}, abs/1711.11561, 2017.

\bibitem{Juliani_consciousfn}
Arthur Juliani, Kai Arulkumaran, Shuntaro Sasai, and Ryota Kanai.
\newblock {On the link between conscious function and general intelligence in
  humans and machines}.
\newblock {\em Transactions on Machine Learning Research}, 2022.
\newblock Survey Certification.

\bibitem{Krizhevsky_ImageNet}
Alex Krizhevsky, Ilya Sutskever, and Geoffrey~E Hinton.
\newblock {ImageNet Classification with Deep Convolutional Neural Networks}.
\newblock In {\em {Advances in Neural Information Processing Systems}},
  volume~25. Curran Associates, Inc., 2012.

\bibitem{kuhl2000new}
Patricia~K Kuhl.
\newblock A new view of language acquisition.
\newblock {\em Proceedings of the National Academy of Sciences},
  97(22):11850--11857, 2000.

\bibitem{Landau_infantshapebias}
Barbara Landau, Linda~B. Smith, and Susan~S. Jones.
\newblock The importance of shape in early lexical learning.
\newblock {\em Cognitive Development}, 3(3):299--321, 1988.

\bibitem{Lee_MetaOptNet}
Kwonjoon Lee, Subhransu Maji, Avinash Ravichandran, and Stefano Soatto.
\newblock {Meta-Learning With Differentiable Convex Optimization}.
\newblock In {\em {{IEEE} Conference on Computer Vision and Pattern Recognition
  ({CVPR})}}, pages 10657--10665. Computer Vision Foundation / {IEEE}, 2019.

\bibitem{Liu_online_self_distillation}
Sihan Liu and Yue Wang.
\newblock {Few-shot Learning with Online Self-Distillation}.
\newblock In {\em {IEEE/CVF} International Conference on Computer Vision
  Workshops ({ICCVW})}, pages 1067--1070. {IEEE}, 2021.

\bibitem{Luo_rectifybgshortcut}
Xu Luo, Longhui Wei, Liangjian Wen, Jinrong Yang, Lingxi Xie, Zenglin Xu, and
  Qi Tian.
\newblock {Rectifying the Shortcut Learning of Background: Shared Object
  Concentration for Few-Shot Image Recognition}.
\newblock {\em CoRR}, abs/2107.07746, 2021.

\bibitem{Marr_shape_importance}
D. Marr, H.~K. Nishihara, and Sydney Brenner.
\newblock Representation and recognition of the spatial organization of
  three-dimensional shapes.
\newblock {\em Proceedings of the Royal Society of London. Series B. Biological
  Sciences}, 200(1140):269--294, 1978.

\bibitem{Oreshkin_TADAM}
Boris~N. Oreshkin, Pau~Rodr{\'{\i}}guez L{\'{o}}pez, and Alexandre Lacoste.
\newblock {{TADAM:} Task dependent adaptive metric for improved few-shot
  learning}.
\newblock In Samy Bengio, Hanna~M. Wallach, Hugo Larochelle, Kristen Grauman,
  Nicol{\`{o}} Cesa{-}Bianchi, and Roman Garnett, editors, {\em {Neural
  Information Processing Systems (NeurIPS)}}, pages 719--729, 2018.

\bibitem{Orhan_humanrobustness}
A.~Emin Orhan and Brenden~M. Lake.
\newblock {Improving the robustness of ImageNet classifiers using elements of
  human visual cognition}.
\newblock {\em CoRR}, abs/1906.08416, 2019.

\bibitem{Qiao_TEWAM}
Limeng Qiao, Yemin Shi, Jia Li, Yonghong Tian, Tiejun Huang, and Yaowei Wang.
\newblock {Transductive Episodic-Wise Adaptive Metric for Few-Shot Learning}.
\newblock In {\em {{IEEE/CVF} International Conference on Computer Vision
  ({ICCV})}}, pages 3602--3611. {IEEE}, 2019.

\bibitem{Sachin_Opt}
Sachin Ravi and Hugo Larochelle.
\newblock {Optimization as a Model for Few-Shot Learning}.
\newblock In {\em International Conference on Learning Representations
  ({ICLR})}. OpenReview.net, 2017.

\bibitem{Ravichandran_Shotfree}
Avinash Ravichandran, Rahul Bhotika, and Stefano Soatto.
\newblock {Few-Shot Learning With Embedded Class Models and Shot-Free Meta
  Training}.
\newblock In {\em {{IEEE/CVF} International Conference on Computer Vision
  (ICCV)}}, pages 331--339. {IEEE}, 2019.

\bibitem{Reddy_humanrobustness}
Manish~V. Reddy, Andrzej Banburski, Nishka Pant, and Tomaso~A. Poggio.
\newblock {Biologically Inspired Mechanisms for Adversarial Robustness}.
\newblock In Hugo Larochelle, Marc'Aurelio Ranzato, Raia Hadsell,
  Maria{-}Florina Balcan, and Hsuan{-}Tien Lin, editors, {\em {Neural
  Information Processing Systems ({NeurIPS})}}, 2020.

\bibitem{Ren_MetaSSFSC}
Mengye Ren, Eleni Triantafillou, Sachin Ravi, Jake Snell, Kevin Swersky,
  Joshua~B. Tenenbaum, Hugo Larochelle, and Richard~S. Zemel.
\newblock {Meta-Learning for Semi-Supervised Few-Shot Classification}.
\newblock {\em CoRR}, abs/1803.00676, 2018.

\bibitem{Ringer_texturebiasfsl}
Sam Ringer, Will Williams, Tom Ash, Remi Francis, and David MacLeod.
\newblock {Texture Bias Of CNNs Limits Few-Shot Classification Performance}.
\newblock {\em CoRR}, abs/1910.08519, 2019.

\bibitem{Rizve_IER_Distill}
Mamshad~Nayeem Rizve, Salman~H. Khan, Fahad~Shahbaz Khan, and Mubarak Shah.
\newblock {Exploring Complementary Strengths of Invariant and Equivariant
  Representations for Few-Shot Learning}.
\newblock In {\em {IEEE} Conference on Computer Vision and Pattern Recognition
  ({CVPR})}, pages 10836--10846. Computer Vision Foundation / {IEEE}, 2021.

\bibitem{Rusu_LEO}
Andrei~A. Rusu, Dushyant Rao, Jakub Sygnowski, Oriol Vinyals, Razvan Pascanu,
  Simon Osindero, and Raia Hadsell.
\newblock {Meta-Learning with Latent Embedding Optimization}.
\newblock In {\em International Conference on Learning Representations
  ({ICLR})}. OpenReview.net, 2019.

\bibitem{Simon_DSNMR}
Christian Simon, Piotr Koniusz, Richard Nock, and Mehrtash Harandi.
\newblock {Adaptive Subspaces for Few-Shot Learning}.
\newblock In {\em {IEEE/CVF} Conference on Computer Vision and Pattern
  Recognition ({CVPR})}, pages 4135--4144. Computer Vision Foundation / {IEEE},
  2020.

\bibitem{Snell_PNFSL}
Jake Snell, Kevin Swersky, and Richard~S. Zemel.
\newblock {Prototypical Networks for Few-shot Learning}.
\newblock In Isabelle Guyon, Ulrike von Luxburg, Samy Bengio, Hanna~M. Wallach,
  Rob Fergus, S.~V.~N. Vishwanathan, and Roman Garnett, editors, {\em Advances
  in Neural Information Processing Systems 30: Annual Conference on Neural
  Information Processing Systems 2017, December 4-9, 2017, Long Beach, CA,
  {USA}}, pages 4077--4087, 2017.

\bibitem{Sobel_Sobel}
Irwin Sobel and Gary Feldman.
\newblock A 3×3 isotropic gradient operator for image processing.
\newblock {\em Pattern Classification and Scene Analysis}, pages 271--272,
  1973.

\bibitem{Song_FSLReview}
Yisheng Song, Ting Wang, Subrota~K. Mondal, and Jyoti~Prakash Sahoo.
\newblock {A Comprehensive Survey of Few-shot Learning: Evolution,
  Applications, Challenges, and Opportunities}.
\newblock {\em CoRR}, abs/2205.06743, 2022.

\bibitem{Lynn_dynamic_obj_recognition}
Lynn K.~A. S{\"o}rensen, Sander~M. Boht{\'e}, Dorina de Jong, Heleen~A.
  Slagter, and H.~Steven Scholte.
\newblock Mechanisms of human dynamic object recognition revealed by sequential
  deep neural networks.
\newblock {\em bioRxiv}, 2022.

\bibitem{Stojanov_3dshapebias}
Stefan Stojanov, Anh Thai, and James~M. Rehg.
\newblock {Using Shape To Categorize: Low-Shot Learning With an Explicit Shape
  Bias}.
\newblock In {\em {IEEE} Conference on Computer Vision and Pattern Recognition
  ({CVPR})}, pages 1798--1808. Computer Vision Foundation / {IEEE}, 2021.

\bibitem{Subramanya_FSLAdversarialRobustness}
Akshayvarun Subramanya and Hamed Pirsiavash.
\newblock A simple approach to adversarial robustness in few-shot image
  classification.
\newblock {\em CoRR}, abs/2204.05432, 2022.

\bibitem{Sung_RelationNet}
Flood Sung, Yongxin Yang, Li Zhang, Tao Xiang, Philip H.~S. Torr, and
  Timothy~M. Hospedales.
\newblock {Learning to Compare: Relation Network for Few-Shot Learning}.
\newblock In {\em {{IEEE} Conference on Computer Vision and Pattern Recognition
  ({CVPR})}}, pages 1199--1208. Computer Vision Foundation / {IEEE} Computer
  Society, 2018.

\bibitem{Taghanaki_sem_seg_survey}
Saeid~Asgari Taghanaki, Kumar Abhishek, Joseph~Paul Cohen, Julien Cohen{-}Adad,
  and Ghassan Hamarneh.
\newblock {Deep semantic segmentation of natural and medical images: a review}.
\newblock {\em {Artificial Intelligence Review}}, 54(1):137--178, 2021.

\bibitem{Tao_poweringfinetuning}
Ran Tao, Han Zhang, Yutong Zheng, and Marios Savvides.
\newblock {Powering Finetuning for Few-shot Learning: Domain-Agnostic Bias
  Reduction with Selected Sampling}.
\newblock {\em {Association for the Advancement of Artificial Intelligence
  (AAAI)}}, 2022.

\bibitem{Tian_RFS}
Yonglong Tian, Yue Wang, Dilip Krishnan, Joshua~B. Tenenbaum, and Phillip
  Isola.
\newblock {Rethinking Few-Shot Image Classification: {A} Good Embedding is All
  You Need?}
\newblock In Andrea Vedaldi, Horst Bischof, Thomas Brox, and Jan{-}Michael
  Frahm, editors, {\em {European Conference in Computer Vision ({ECCV})}},
  volume 12359 of {\em Lecture Notes in Computer Science}, pages 266--282.
  Springer, 2020.

\bibitem{Vinyals_MatchingNets}
Oriol Vinyals, Charles Blundell, Tim Lillicrap, Koray Kavukcuoglu, and Daan
  Wierstra.
\newblock {Matching Networks for One Shot Learning}.
\newblock In Daniel~D. Lee, Masashi Sugiyama, Ulrike von Luxburg, Isabelle
  Guyon, and Roman Garnett, editors, {\em Advances in Neural Information
  Processing Systems 29}, pages 3630--3638, 2016.

\bibitem{Wang_FSLReview}
Yaqing Wang, Quanming Yao, James~T. Kwok, and Lionel~M. Ni.
\newblock {Generalizing from a Few Examples: {A} Survey on Few-shot Learning}.
\newblock {\em {ACM} Comput. Surv.}, 53(3):63:1--63:34, 2020.

\end{thebibliography}
}

\clearpage
\newpage
\appendix
\section{Appendix}


\subsection{Methodology}
\label{section:app_methodology}
The algorithm for the proposed LSFSL approach is provided in Algorithm~\ref{algo:lsfsl_simple}. 
The sequential distillation and online self-distillation approach to incorporate shape information is illustrated in Algorithm~\ref{algo:lsfsl_distill} and Algorithm~\ref{algo:lsfsl_online}.


\begin{algorithm}[!htbp]
\centering
\caption{LSFSL-Distill: Training Algorithm}
\label{algo:lsfsl_distill}
\begin{algorithmic}[1]
 \Statex \textbf{Input:} dataset $\mathcal{D}$, fixed LSFSL-trained RIN model $R_t$ with feature extractor $f_t$ and classifier $g_t$, randomly initialized student RIN model $R_s$ with feature extractor $f_t$ and classifier $g_t$, epochs $E$, softmax operator $\sigma$, cross-entropy loss $CE$, Kullback-Leibler divergence loss $KLD$, cross-entropy loss factor $\alpha$, teacher-student decision alignment loss factor $\beta$
\For {epoch $e \in \{1, 2,..,E\}$}
  \State sample a mini-batch ${(x, y)} \sim \mathcal{D}$
  \State $R_t(x)$ = $g_{t} (f_{t}(x))$
  \State $R_s(x)$ = $g_{s} (f_{s}(x))$
  \State $\mathcal{L}_{CER}$ = $CE(R_s(x), y)$
  \State $\mathcal{L}_{DA}$ = $KLD(R_t(x), R_s(x))$
  \State $\mathcal{L}$ = $\alpha \mathcal{L}_{CER}$ + $\beta \mathcal{L}_{DA}$
 \State Update parameters of $R_s$ based on $\mathcal{L}$ using Stochastic Gradient Descent (SGD)
\EndFor
\State \Return{RIN student model $R_s$}
\end{algorithmic}
\end{algorithm}


\begin{algorithm}[!t]
\centering
\caption{LSFSL-Online: Training Algorithm}
\label{algo:lsfsl_online}
\begin{algorithmic}[1]
 \Statex \textbf{Input:} 
 dataset $\mathcal{D}$, randomly initialized RIN model $R$ with feature extractor $f_{\Phi}$ and classifier $g_{\Theta}$, randomly initialized SIN model $S$ with feature extractor $f_{\phi}$ and classifier $g_{\omega}$, randomly initialized and fixed teacher RIN model $R_t$ with feature extractor $f_{t, \Phi}$ and classifier $g_{t, \Theta}$, epochs $E$, softmax operator $\sigma$, stopgrad operator $SG$, cross-entropy loss $CE$, Kullback-Leibler divergence loss $KLD$, mean square error $MSE$, Sobel edge operator $h$, feature alignment loss factors ($\gamma_r$, $\gamma_r$) , decision alignment loss factors ($\lambda_r$, $\lambda_s$), teacher-student decision alignment loss factor $\beta$ 
\For {epoch $e \in \{1, 2,..,E\}$}
  \State sample a mini-batch ${(x, y)} \sim \mathcal{D}$
  \State $x_{shape} = h(x)$
   \State $z_{\Phi}$ = $f_{\Phi}(x)$
  \State $z_{\phi}$ = $f_{\phi}(x_{shape})$
  \State $R(x)$ = $g_{\Theta} (f_{\Phi}(x))$
  \State $S(h(x))$ = $g_{\omega} (f_{\phi}(h(x_{shape})))$
  \State $R_t(x)$ = $g_{t,\Theta} (f_{t,\Phi}(x))$
  \State $\mathcal{L}_{CER}$ = $CE(\sigma(R(x)), y)$
  \State $\mathcal{L}_{CES}$ = $CE(\sigma(S(h(x))), y)$
  \State $\mathcal{L}_{FAR}$ = $MSE(z_{\Phi}, SG(z_{\phi}))$ \Comment{\small (Eq. \ref{equ:fa_rin})}
  \State $\mathcal{L}_{FAS}$ = $MSE(SG(z_{\Phi}), z_{\phi})$ \Comment{\small (Eq. \ref{equ:fa_sin})}
  \State $\mathcal{L_{FA}}$ = $\gamma_r \mathcal{L}_{FAR}$ + $\gamma_s \mathcal{L}_{FAS}$ \Comment{\small (Eq. \ref{equ:fa_mse})}
   \State $\mathcal{L}_{DAR}$ = $KLD(\sigma(R(x)))$, $SG(\sigma(S(h(x))))$ \Comment{\small (Eq. \ref{equ:da_rin})}
   \State $\mathcal{L}_{DAS}$ = $KLD(SG(\sigma(R(x)))$, $\sigma(S(h(x))))$ \Comment{\small (Eq. \ref{equ:da_sin})}
  \State $\mathcal{L_{DA}}$ = $\lambda_r \mathcal{L}_{DAR}$ + $\lambda_s \mathcal{L}_{DAS}$ \Comment{\small (Eq. \ref{equ:da_kl})}   
   \State $\mathcal{L}_{TS}$ = $KLD(\sigma(R(x))$, $\sigma(R_t(x)))$
  \State $\mathcal{L}$ = $\mathcal{L}_{CER}$ + $\mathcal{L}_{CES}$ + $\mathcal{L}_{FA}$ + $\mathcal{L}_{DA}$ + $\beta \mathcal{L}_{TS}$
  \State Update parameters of $R$ and $S$ based on $\mathcal{L}$ using Stochastic Gradient Descent
  \State Update $R_t$ as EMA of $R$
\EndFor
\State \Return{RIN student model $R$}
\end{algorithmic}
\end{algorithm}


\subsection{Analysis Visualizations}
\label{section:app_visualizations}

\begin{figure}[!t]
\centering
\includegraphics[width=\linewidth]{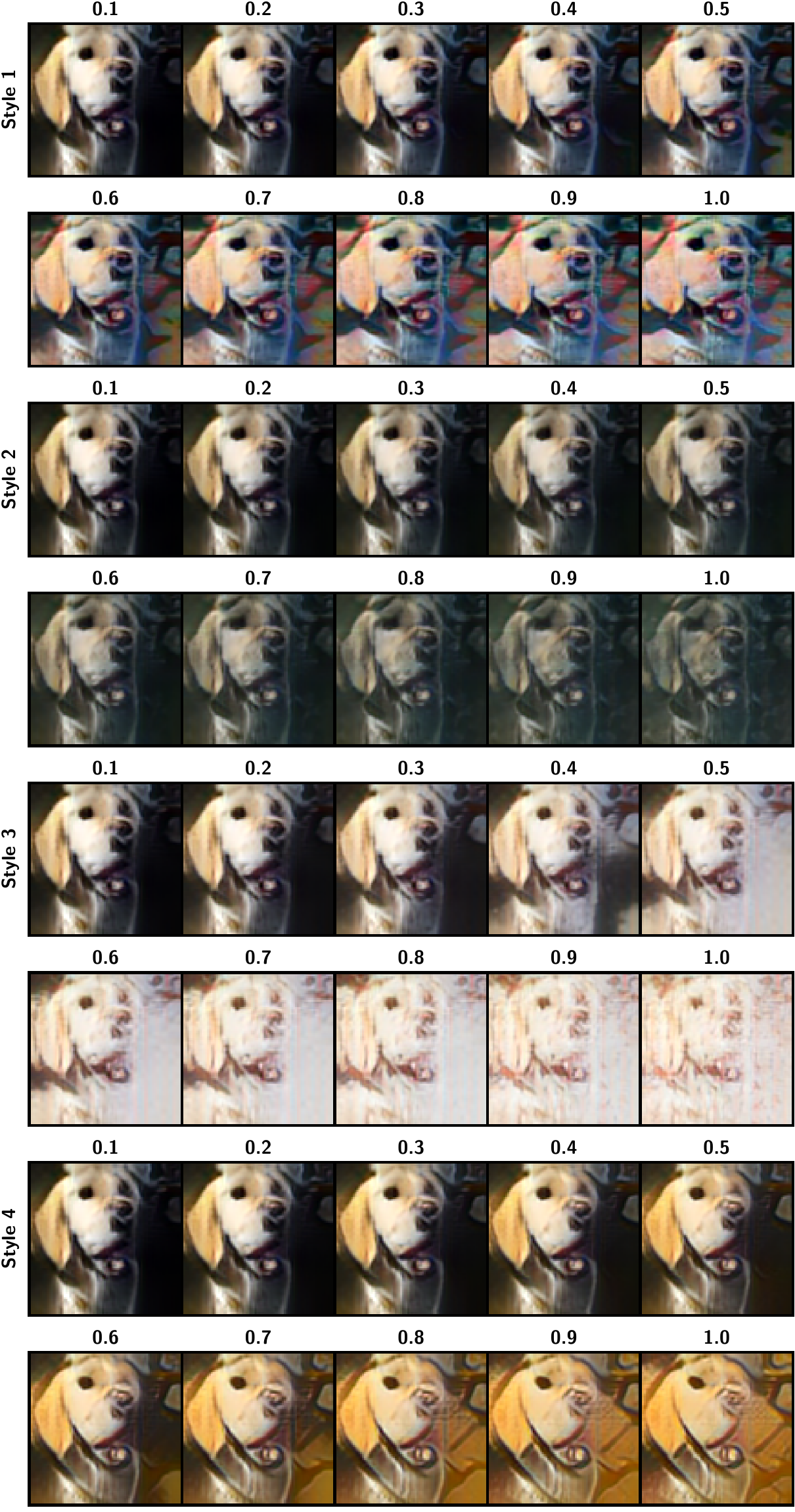}
\caption{An illustration of the various stylized miniImageNet images generated for varying stylization intensities to perform the texture bias analysis in Section~\ref{section:texture_bias}.}
\label{fig:stylize}
\end{figure}
\begin{figure}[!t]
\centering
\includegraphics[width=\linewidth]{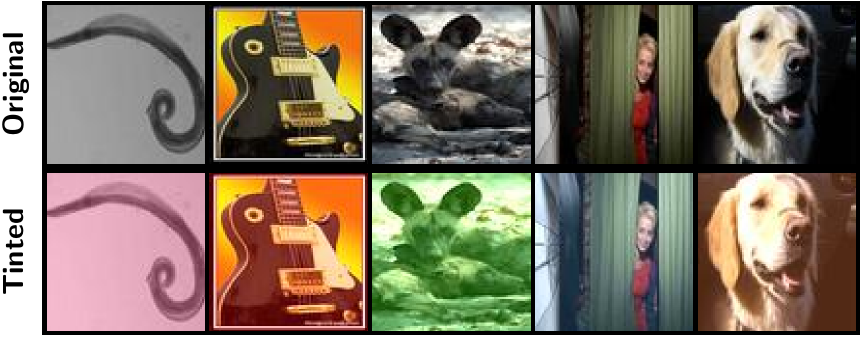}
\caption{An illustration of class-specific tinted images generated for spurious correlation analysis is provided in Section~\ref{section:spurious_correlation_analysis}.}
\label{fig:tint}
\end{figure}
\begin{figure}[!t]
\centering
\includegraphics[width=\linewidth]{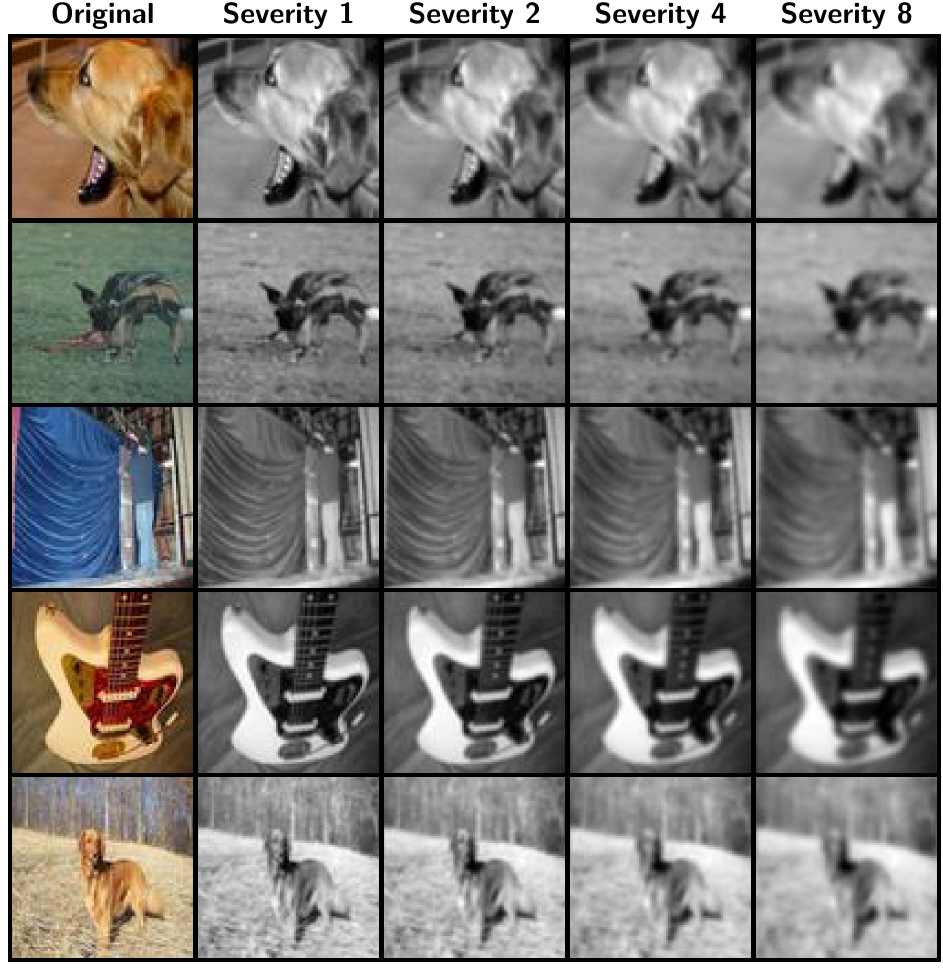}
\caption{An illustration of low-pass filtered images generated for statistical regularity analysis in Section~\ref{section:statistical_regularity_analysis}.}
\label{fig:ft}
\end{figure}

The texture bias, spurious correlation, and statistical regularity analysis are performed by applying different textures by stylization, class-specific tints, and radial low-pass Fourier filters at increasing severities as shown in Figure~\ref{fig:stylize}, Figure~\ref{fig:tint} and Figure ~\ref{fig:ft}.

\end{document}